\documentclass[conference]{IEEEtran}
\IEEEoverridecommandlockouts

\usepackage{cite}
\usepackage{amsmath,amssymb,amsfonts}
\usepackage{algorithmic}
\usepackage{graphicx}
\usepackage{textcomp}
\usepackage{xcolor}
\usepackage{multirow}
\usepackage[utf8]{inputenc} 
\usepackage[T1]{fontenc}    
\usepackage{mathptmx}       

\def\BibTeX{{\rm B\kern-.05em{\sc i\kern-.025em b}\kern-.08em
    T\kern-.1667em\lower.7ex\hbox{E}\kern-.125emX}}
\begin{document}

\title{Enhancing Whisper's Accuracy and Speed \\ for Indian Languages through \\ Prompt-Tuning and Tokenization

}

\author{\IEEEauthorblockN{Kumud Tripathi, Raj Gothi, Pankaj Wasnik} 
\IEEEauthorblockA{
Media Analysis Group, Sony Research India\\
\{kumud.tripathi, raj.gothi, pankaj.wasnik\}@sony.com}}

\maketitle

\begin{abstract}
Automatic speech recognition has recently seen a significant advancement with large foundational models such as Whisper. However, these models often struggle to perform well in low-resource languages, such as Indian languages. This paper explores two novel approaches to enhance Whisper's multilingual speech recognition performance in Indian languages. First, we propose prompt-tuning with language family information, which enhances Whisper's accuracy in linguistically similar languages. Second, we introduce a novel tokenizer that reduces the number of generated tokens, thereby accelerating Whisper's inference speed. Our extensive experiments demonstrate that the tokenizer significantly reduces inference time, while prompt-tuning enhances accuracy across various Whisper model sizes, including Small, Medium, and Large. Together, these techniques achieve a balance between optimal WER and inference speed.


\end{abstract}

\begin{IEEEkeywords}
Multilingual speech recognition, whisper, prompt-tuning, tokenizer, inference speed
\end{IEEEkeywords}

\section{Introduction}
Automatic Speech Recognition (ASR) systems have transformed human-machine interaction, enabling voice-based access to information across various domains \cite{sayers2021dawn}. This is particularly important in India, where over 300 million people cannot read and more than 60 languages are spoken. Developing reliable ASR systems is essential to serve the diverse Indian population effectively \cite{census2011}. However, the linguistic diversity and limited representation of Indian languages in publicly available global datasets pose substantial challenges even for state-of-the-art (SOTA) ASR models such as Whisper \cite{dey2022overview}. 

Unlike monolingual approaches, Multilingual Speech Recognition (MSR) leverages shared learning across languages, improving overall accuracy by utilizing linguistic similarities \cite{pmlr-v202-radford23a, JMLR:v25:23-1318, zhang2023google}. Modified Whisper models for Indian languages address these challenges by incorporating techniques like prompting to enhance recognition accuracy \cite{10446990}. Despite these advancements, Whisper's effectiveness in Indian languages is hampered by deficiencies in tokenization. The tokenization process, which is crucial for ASR speed, affects low-resource languages more heavily \cite{li2024improving}. High-resource languages benefit from extensive token sets, whereas low-resource languages face slower inference times due to fewer tokens in the pre-trained Whisper tokenizer. 

This research proposes two innovative strategies to enhance the efficiency of the Whisper model for Indian languages. We utilize prompt-tuning with language family information to reduce Word Error Rate (WER) by addressing phonetic and linguistic similarities. Additionally, we introduce a customized tokenizer for Indian languages to improve the Whisper's efficiency during the inference time. Both approaches individually surpass baseline ASR model performance and when combined, they achieve a balanced trade-off between accuracy and inference time. Our research aims to significantly improve the accuracy and applicability of ASR systems across diverse Indian languages, advancing the development of a more effective and reliable MSR model.

\section{Related Work}
In Indian language ASR, recent studies show that fine-tuned large pre-trained models outperform those trained from scratch \cite{bhogale23_interspeech,javed2021building,gupta2021clsril}. This approach leverages the extensive knowledge and features already embedded in pre-trained models to fine-tune them using the data from specific languages to enhance performance. We have used OpenAI's Whisper model, representing a significant advancement in ASR technology through its multitasking and multilingual training approach on weakly supervised data \cite{pmlr-v202-radford23a}. This methodology allows the model to leverage shared learning across multiple languages, enhancing its overall accuracy and robustness. However, its performance in Indian languages, characterized by their linguistic diversity and low representation in the training dataset, poses challenges and yields a high WER. 

Unlike Whisper's multilingual approach, the IndicWhisper paper \cite{bhogale23_interspeech}, focuses on fine-tuning models individually for each Indian language without altering the architecture. While this method achieves good performance, it lacks the scalability and efficiency of a multilingual model that can handle multiple languages simultaneously. 
A recent survey \cite{yadav-sitaram-2022-survey} emphasizes the benefits of training ASR models based on language families, which can enhance performance by leveraging shared phonetic and linguistic features.
Task-specific prompt fine-tuning in large language models (LLMs) has demonstrated considerable benefits by sharing knowledge across tasks \cite{minaee2024large}. 
Similarly, prompting techniques in ASR, such as targeting the speaker \cite{ts_whisper} or leveraging different prompt techniques, can significantly enhance performance in multilingual settings \cite{10446990}.

Existing research has explored various methods to enhance ASR performance in Indian languages, including leveraging language identification (ID) as a prompt to the decoder model \cite{jayakumar23_interspeech}. However, this approach has already been integrated into Whisper’s decoder model. Another approach involves common label mapping, where different language characters are mapped to a common set of labels based on phonetic similarities \cite{verma23_interspeech}. However, these methods typically involve model training from scratch and do not fully capitalize on the advantages of pre-trained supervised training. They also often require additional modules to resolve conflicts from the same label representing different characters in different languages. Thus, to enhance and study the effectiveness of prompting, we proposed incorporating language family-based prompts alongside language ID for the Whisper multilingual model.
 
Stemming our motivation from the recent NLP research indicating that enhancing tokenizers with language-specific tokens and training on substantial language-specific datasets improves performance, especially for low-resource languages \cite{hangya-etal-2022-improving}. We anticipate that this principle also applies to the Whisper, suggesting that expanding token sets with diverse textual data can boost ASR accuracy, particularly in low-resource language scenarios. However, for low-resource languages, such as Indian languages, tokenization presents challenges due to the limited token sets. We demonstrated that customizing tokenization methods tailored to these languages enhances inference efficiency and overall ASR performance.



 
\section{Dataset}

For our study, we utilized a diverse set of 8 languages representing two major language families of the Indian subcontinent:
\begin{itemize}
    \item \textbf{Indo-Aryan Languages}: Hindi, Gujarati, Marathi, Bengali
    \item \textbf{Dravidian Languages}: Tamil, Telugu, Kannada, Malayalam
\end{itemize}
\noindent These languages were selected to ensure broad coverage and representation across Indo-Aryan and Dravidian language groups, with four languages from each family. To this extent, our study utilized the Kathbath dataset \cite{javed2023indicsuperb} which comprises read speech data collected for 12 Indian languages, including the aforementioned eight languages. The dataset is sourced from Wikipedia articles and news sources, making it a comprehensive resource for speech research. Kathbath supports a variety of speech tasks, notably Automatic Speech Recognition, Speaker Verification, and Language Identification. It serves as the foundational dataset for the IndicSUPERB benchmark, dedicated to enhancing speech language understanding specifically for Indic languages. In our study, Kathbath was utilized primarily for ASR. We used the same training, validation, and testing datasets as discussed in the paper \cite{javed2023indicsuperb}. 

\begin{figure}[t]
\centering
\includegraphics[width=1\columnwidth]{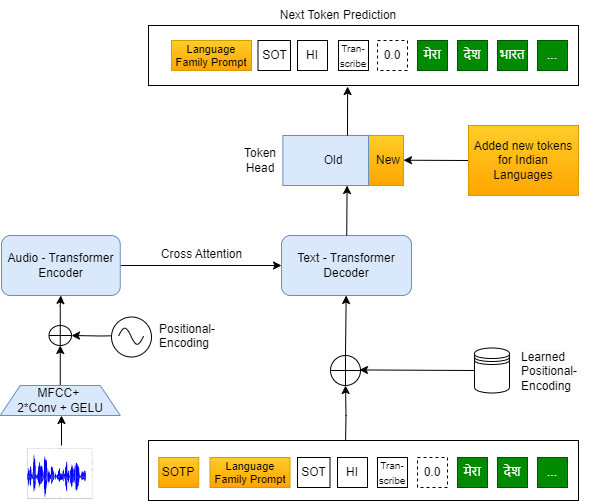} 
\caption{Block diagram of the proposed architecture. Yellow color represents the proposed approaches.}
\label{architecture}
\end{figure}

\section{Proposed Methodology}
This section provides an overview of the Whisper model and introduces our two proposed techniques for fine-tuning it. The first technique involves language family-based prompting and the second one introduces a new tokenizer for Indian languages.


\subsection{Whisper Model}
Whisper is a multitask and multilingual model for speech-related tasks. It has transformer encoder-decoder-based architecture as shown in Figure \ref{architecture}. Here, the encoder learns audio representation, and the decoder learns the task-specific aspects. The model takes the log-mel spectrograms as input and passes them to the convolution module, followed by the transformer encoder. At the decoder side, it takes the Start token \textless SOT\textgreater{}, Language ID token, Task-specific token, and No Time-stamp token (If time-stamp is not required) followed by predicted tokens. \textless Transcribe\textgreater{} and \textless Translation\textgreater{} are the two task-based tokens passed to the decoder model. For our work, we are interested in speech recognition; hence, we set the decoder for the \textless Transcribe\textgreater{} task. 
During inference, the decoder model predicts each token iteratively, appending the newly predicted token to the decoder side as input.

\begin{table*}[t]
\caption{WER (in \%) and inference time (in min.) on Kathbath using Whisper Medium-based baseline and proposed models. } \label{tab:main} 
\resizebox{\textwidth}{!}
{\begin{tabular}{|l|l|l|l|l|l|l|l|l|l|l|l|l|} \hline
\multirow{4}{*}{\textbf{Languages}} & \multicolumn{6}{|c|}{\textbf{WER (in \%)}} & \multicolumn{6}{|c|}{\textbf{Inference Time (in Minutes)}} \\ \cline{2-13}
 & \multicolumn{3}{|c|}{\textbf{Baseline}} &\multicolumn{3}{|c|}{\textbf{Proposed}} & \multicolumn{3}{|c|}{\textbf{Baseline}} &\multicolumn{3}{|c|}{\textbf{Proposed}} \\ \cline{2-13}
& W-M PT & \begin{tabular}[c]{@{}l@{}}Indic\\Whisper\end{tabular} & W-M FT &  \begin{tabular}[c]{@{}l@{}}W-M FT \\ w/ Pro\end{tabular} & \begin{tabular}[c]{@{}l@{}}W-M FT \\ w/ Tok\end{tabular} & \begin{tabular}[c]{@{}l@{}}W-M FT w/ \\ Pro + Tok\end{tabular} &W-M PT & \begin{tabular}[c]{@{}l@{}}Indic\\Whisper\end{tabular} & W-M FT &  \begin{tabular}[c]{@{}l@{}}W-M FT \\ w/ Pro\end{tabular} & \begin{tabular}[c]{@{}l@{}}W-M FT \\ w/ Tok\end{tabular} & \begin{tabular}[c]{@{}l@{}}W-M FT w/ \\Pro + Tok\end{tabular} \\ \hline
Hindi & 42.64& 10.09& 11.24& 10.31& 11.10& 11.48& 29.88&43.92 & 27.95& 45.63& 19.70& 32.46\\
Gujarati & 113.45& 17.81& 16.69& 15.01& 15.78& 16.55& 33.55& 99.66& 64.51& 103.17& 21.30& 35.02\\
Marathi & 106.68& 19.83& 16.44& 14.80& 15.94& 16.29& 43.72& 49.59& 31.52& 52.85& 23.46& 38.04\\
Bengali & 134.98& 16.65& 13.22& 11.78& 12.69& 12.55& 121.05& 85.48& 53.82& 90.96& 24.48& 40.38\\ \hline
Tamil & 63.90& 24.22& 24.51& 23.08& 24.54& 24.90& 31.02& 53.50& 31.40& 55.59& 29.01& 47.21\\
Telugu & 138.19& 25.01& 23.68& 22.12& 23.69& 23.81& 90.58& 79.17& 48.35& 85.79& 23.84& 40.77\\
Kannada & 104.55& 19.33& 19.45& 17.98& 18.49& 18.72& 81.42& 69.30& 42.18& 73.88& 22.13& 38.25\\
Malayalam & 134.40& 34.81& 35.99& 33.23& 35.15& 35.74& 100.29& 102.13& 61.51& 98.57& 26.12& 42.69\\\hline
Average & 104.84& 20.96& 20.15& \textbf{18.53}& 19.67& 20.05& 66.43& 72.84& 45.15& 75.81& \textbf{23.75}& 39.35 \\\hline
\end{tabular}}
\end{table*}

\subsection{Prompting Whisper}
We explored different fine-tuning techniques in Whisper by providing language-family-based prompts to the decoder side. As shown in Figure \ref{architecture}, we added two extra tokens before the \textless SOT\textgreater{} token: \textless SOTP\textgreater{} and a language family-based prompt during both fine-tuning and inference. The \textless SOTP\textgreater{} token is already available in Whisper architecture and is used with a prompt for generating conditional text during inference. During fine-tuning, languages from the same family are given the same prompt. For example, Hindi, Gujarati, Marathi, and Bengali share the same language family-based prompt because they belong to the Indo-Aryan language family. Similarly, the Dravidian languages share a common prompt. This ensures that the fine-tuning process incorporates linguistic similarities within language families.

\subsection{Tokenizer}
A tokenizer breaks text into smaller units called tokens, essential for processing by models like GPT-2 \cite{radford2019language}. Each token represents a text fragment, enabling the model to understand language. In Whisper, the GPT-2 tokenizer is used directly for speech tasks. Encoding converts raw text into a format the model can process, while decoding translates the model's output back into readable text. Whisper's tokenizer efficiently handles iterative predictions and plays a critical role in both processes. Indian languages have few tokens in the existing Whisper tokenizer, so additional Byte Pair Encoding (BPE) tokens were incorporated, derived from datasets emphasizing common sequences in Indian languages. As new BPE tokens were added, Whisper's architecture required adjustments, specifically modifying the last layer (token head) to include new random weights for the added tokens while retaining original weights to preserve prior learning, as shown in Figure \ref{architecture}. The token head's dimension was expanded, and a softmax function was applied during prediction to select the token with the highest probability, ensuring coherent language generation.

\section{Experiment and Results}

\subsubsection*{\textbf{Experimental Setup}}
We conducted our experiments using 4 NVIDIA A100 (40GB) GPUs. During the experiments, we fine-tuned the Whisper model (W) in 3 different sizes, denoted by W-$m$, where $m=\{S, M, L\}$ for small, medium and large size, respectively.
All the proposed prompting-based and tokenizer-based Whisper models are fine-tuned for five epochs and three epochs for all the remaining models. We used a learning rate of $1\times$10$^{-5}$ and four gradient accumulation steps. We report numbers on eight Indian languages taken from the Kathbath dataset that is publicly available.

Firstly, we evaluated the pre-trained Whisper model (W-$m$ PT) with different sizes indicated by $m$ \cite{pmlr-v202-radford23a}. Additionally, we considered the {IndicWhisper} as one of our baseline. 
To compare our techniques, we also fine-tuned the Whisper models in a multilingual context with and without the proposed techniques, denoted as W-$m$ FT.



We further fine-tuned the Whisper model using language family-specific prompts to leverage the language family information. We used the prompt \textless indo\textgreater{} for Indo-Aryan languages and \textless dra\textgreater{} for Dravidian languages. Additionally, we introduced new BPE tokens for each language into the existing Whisper tokenizer and fine-tuned the model in a multilingual setting, referred to as W-m FT ($y$), where $y$ is the number of added tokens per language.
We also combined both techniques: prompt-tuning and the new tokenizer ($y=$ 250) and labeled this variant as W-$m$ FT with prompt+tokenizer. Best results in the tables are highlighted in bold.

\subsubsection*{\textbf{Result Analysis}}
From Table \ref{tab:main}, we can see that the proposed models outperform the baseline models in both WER and inference time. Among the baseline models, W-M FT surpasses W-M PT and IndicWhisper in WER, particularly for Gujarati, Marathi, Bengali, and Telugu. In the proposed models, W-M FT with language family prompts (Pro) achieves the best WER, outperforming both W-M FT with only the tokenizer (Tok) and W-M FT with both prompt and tokenizer, achieving lower WERs across all languages.
In terms of inference time, W-M FT among the baseline models outperforms W-M PT and IndicWhisper for all languages. Among the proposed models, W-M FT with the tokenizer achieves the fastest inference time, outperforming both W-M FT with prompt and W-M FT with both prompt and tokenizer.
Thus, W-M FT with prompts provides the best WER, W-M FT with the tokenizer offers the fastest inference, and W-M FT with both prompt and tokenizer balances optimal WER and inference time.


\begin{table}[h]
\centering
\caption{WER (in \%) on Kathbath using Whisper Medium with our Tokenizer on various numbers of tokens ($y$) added per language.} \label{tab:token} 
\small
\resizebox{0.8\linewidth}{!}
{\begin{tabular}{|l|l|l|l|l|} \hline
\multirow{1}{*}{{\textbf{Languages}}} & \begin{tabular}[c]{@{}l@{}}\textbf{W-M FT}\\\textbf{(y=1000)}\end{tabular} & \begin{tabular}[c]{@{}l@{}}\textbf{W-M FT}\\\textbf{(y=500)}\end{tabular} & \begin{tabular}[c]{@{}l@{}}\textbf{W-M FT}\\\textbf{(y=250)}\end{tabular} & \begin{tabular}[c]{@{}l@{}}\textbf{W-M FT}\\\textbf{(y=125)}\end{tabular} \\ \hline
Hindi & 12.53 & 11.40 & \textbf{11.10} & 11.82 \\
Gujarati & 17.92 & 16.30 & \textbf{15.78} & 18.09 \\
Marathi & 17.14 & 16.26 & \textbf{15.94 }& 16.92 \\
Bengali & 13.94 & 12.58 & \textbf{12.69} & 13.99 \\\hline
Tamil & 25.66 & 24.98 & \textbf{24.54} & 25.56 \\
Telugu & 25.99 & 24.17 & \textbf{23.69} & 24.41 \\
Kannada & 21.03 & 18.99 & \textbf{18.49} & 21.11 \\
Malayalam & 37.69 & 36.94 & \textbf{35.15} & 37.13 \\\hline
Average & 21.48 & 20.20 & \textbf{19.67} & 21.12\\\hline
\end{tabular}}
\end{table}

In Table \ref{tab:token}, we present the impact of adding different numbers of additional tokens to the existing Whisper tokenizer. The results reveal that the W-M FT (y=250) configuration outperforms other tokenized models, suggesting that adding 250 additional tokens represents an optimal threshold that enhances the model's performance.


\begin{table}[h]
\caption{Number of generated tokens with and without our Tokenizer.} \label{tab:token1} 
\small
\resizebox{\linewidth}{!}
{\begin{tabular}{|l|l|l|l|} \hline
\multirow{2}{*}{\textbf{Languages}} & \multirow{2}{*}{\textbf{Sentences}} & \multicolumn{2}{|c|}{\textbf{Generated Tokens}} \\ \cline{3-4} 
 &  & w/o Tok & w/ Tok \\ \hline
English & I love my country & 4 Tokens & 4 Tokens \\ \hline
Hindi & \includegraphics[width=0.4\linewidth]{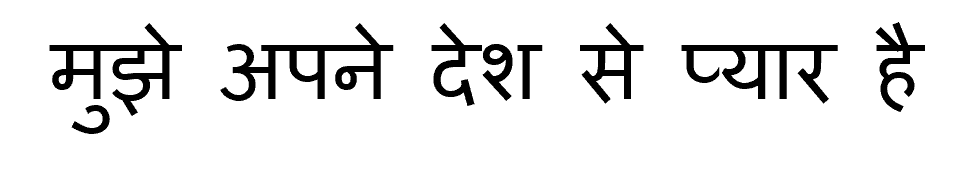} & 27 Tokens & 19 Tokens \\\hline
Malayalam & \includegraphics[width=0.4\linewidth]{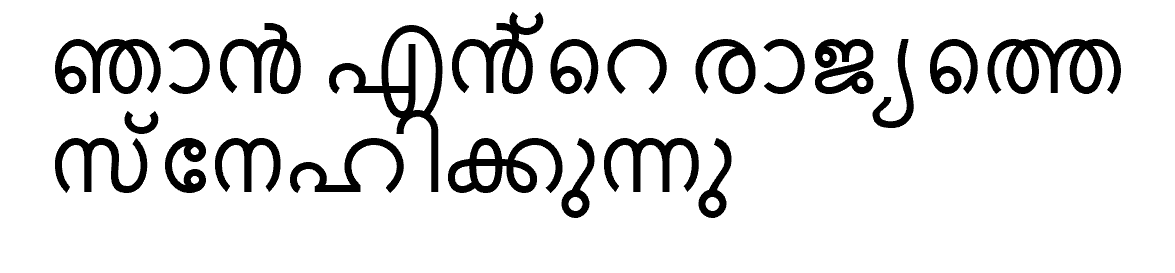} & 79 Tokens & 31 Tokens \\ \hline
\end{tabular}}
\end{table}

\begin{table}[t]
\caption{Inference time (in min.) and WER (in \%) for various Whisper models with and without our proposed Tokenizer ($y$=250).} \label{tab:time} 
\small
\resizebox{\linewidth}{!}
{\begin{tabular}{|l|l|l|l|l|l|l|} \hline
\multirow{3}{*}{\textbf{Languages}} & \multicolumn{2}{|c|}{\textbf{W-S FT}} & \multicolumn{2}{|c|}{\textbf{WhisperX}} & \multicolumn{2}{|c|}{\textbf{Faster Whisper}} \\ \cline{2-7}
  & w/o Tok & w/ Tok & w/o Tok & w/ Tok & w/o Tok &  w/ Tok \\ \hline \hline
 \multicolumn{7}{|c|}{\textbf{Inference Time (in Minutes)}} \\ \hline \hline
Hindi & 14.71 & 14.60 &36.15 & 26.34 &27.65 &20.49  \\
Gujarati & 33.84 & 14.37 &37.84 &26.72 &30.07  &21.61  \\
Marathi & 16.59 & 12.15 &39.53 &28.54 &31.24 &23.09  \\
Bengali & 28.69 & 16.08 &39.72 &29.72 &35.87 &23.84  \\\hline
Tamil & 16.20 & 15.81 &39.64 &32.97 &35.93 &29.55  \\
Telugu & 26.37 & 17.42 &42.67 & 28.22 &39.67 &25.24  \\
Kannada & 23.02 & 15.54  &54.67 & 25.78 & 51.61 &23.53  \\
Malayalam & 31.74 & 17.73 &32.57 &28.61 &30.05 &25.26  \\\hline
Average & 23.91 & \textbf{15.46} &40.34 &\textbf{28.36} & 35.26  &\textbf{24.07} \\\hline \hline
 \multicolumn{7}{|c|}{\textbf{WER (in \%)}} \\ \hline \hline
Hindi & 14.44 & 13.80 & 37.99 & 10.91 & 40.88 & 18.74 \\
Gujarati & 20.55 & 18.21 & 109.54 & 14.99 & 111.54 & 26.22\\
Marathi & 20.65 & 22.69 & 88.64 & 14.78 & 88.54 & 24.11\\
Bengali & 17.84 & 18.62 & 103.41 & 14.08 & 107.82 & 22.09\\ \hline
Tamil & 28.56 & 26.99 & 59.24 & 22.31 & 63.82 & 32.73\\
Telugu & 28.24& 26.71 & 101.29 & 22.56 & 111.35 & 31.09\\
Kannada & 24.27 & 21.10 & 96.33 & 18.60 & 99.61 & 27.29\\
Malayalam & 41.79 & 42.37 & 100.08 & 33.69 & 120.02 & 44.46\\\hline
Average & 24.54 & \textbf{23.81} & 87.06 & \textbf{18.99} & 92.94 & \textbf{28.34}\\\hline
\end{tabular}}
\end{table}

\begin{table}[hbt!]
\caption{WER (in \%) on Kathbath using various Whisper
models with and without our proposed Prompt.}\label{tab:prompt1} 
\centering
\small
\resizebox{0.9\linewidth}{!}
{\begin{tabular}{|l|l|l|l|l|} \hline
\multirow{2}{*}{\textbf{Languages}} & \multicolumn{2}{|c|}{\textbf{W-S FT}} & \multicolumn{2}{|c|}{\textbf{W-L FT}} \\ \cline{2-5}
 & w/o Prompt& w/ Prompt& w/o Prompt & w/ Prompt\\ \hline
Hindi & 14.44 &13.81 &9.56 & 9.24 \\
Gujarati & 20.55 &18.19 &14.57 & 13.95 \\
Marathi & 20.65 &22.69 &14.21 & 13.47 \\
Bengali & 17.84 &18.61 &11.34 & 10.30 \\\hline
Tamil & 28.56 & 27.01 & 22.80 & 21.85 \\
Telugu & 28.24 & 26.71 &21.69 & 20.34 \\
Kannada & 24.27 & 21.10 & 17.13 & 16.48 \\
Malayalam & 41.79 & 42.37 & 34.02 & 31.63 \\\hline
Average & 24.54 & \textbf{23.81} &18.16 & \textbf{17.15}\\\hline
\end{tabular}}
\end{table}


Table \ref{tab:token1} compares the number of tokens generated by the original Whisper tokenizer (column 3) with our new tokenizer (column 4) for various languages. English sentences serve as a baseline with 4 tokens. For Hindi and Malayalam, our tokenizer significantly reduces token count: from 27 to 19 tokens in Hindi and from 79 to 31 tokens in Malayalam. These reductions demonstrate that our tokenizer effectively minimizes the number of tokens, thereby enhancing decoder inference time efficiency. Since the decoder model operates auto-regressively, relying on previously generated tokens, reducing the token count leads to faster and more efficient processing for Indian languages.


In Table \ref{tab:time}, we present a detailed comparison of inference times and WER between existing Whisper models with and without our new tokenizer. The results clearly show a significant improvement in inference speed across all variants when using our tokenizer. This enhancement is particularly notable compared to SOTA models designed for fast inference, such as WhisperX \cite{bain2023whisperx} and Faster Whisper \cite{fw}. 
Our tokenizer reduces processing time by minimizing the number of tokens generated for each language, making it crucial for real-time applications. 
Additionally, Table \ref{tab:time} shows that our tokenizer consistently improves WER performance across all Whisper model variants. Overall, the result confirms that our tokenizer not only reduces inference time but also enhances performance, regardless of the model's size.



In Table \ref{tab:prompt1}, we conducted an ablation study focused on the impact of prompt fine-tuning across various Whisper model sizes, i.e., the Small and Large versions. 
\textbf{The results demonstrate that prompt fine-tuning generally
improves performance across most model sizes and languages. However, its effectiveness can vary due to language-specific challenges. }
This study highlights the versatility of prompt fine-tuning, making even the smaller models more competitive with larger ones in terms of accuracy and efficiency.

\section{Conclusion}
In this work, we demonstrate a significant advancement in multilingual speech recognition for Indian languages using the Whisper model. We have successfully improved the model accuracy for underrepresented Indian languages. By incorporating prompt-tuning with language family information, we leveraged linguistically related languages. Additionally, we introduced a new tokenizer to enhance the model's efficiency in terms of inference time by reducing the number of generated tokens without compromising performance. Our consistently experiments show that both prompt fine-tuning and the proposed tokenizer individually outperform baseline ASR models, and their combination achieves an optimal balance between WER and inference speed. The resulting efficient Whisper model provides a flexible solution, enabling users to adjust the trade-off between accuracy and speed according to their specific application needs. In future work, we aim to fine-tune the model on more low-resource languages and dialects to enhance performance in diverse linguistic settings.

\bibliographystyle{IEEEtran}
\bibliography{IEEEabrv,IEEEtran}


\end{document}